\documentclass[journal]{IEEEtran}

\usepackage{amsmath,amssymb,amsfonts}
\usepackage{algorithm}
\usepackage{algpseudocode}
\usepackage{array}
\usepackage{textcomp}
\usepackage{stfloats}
\usepackage{url}
\usepackage{verbatim}
\usepackage{graphicx}
\usepackage{cite}
\usepackage{xcolor}
\usepackage{float}

\usepackage{caption}
\usepackage{subcaption} 

\usepackage{tikz}
\usepackage{tikz-qtree}
\usetikzlibrary{shapes.geometric, arrows, arrows.meta, positioning, calc}
\usepackage{listings}
\usepackage{tabularx}
\usepackage{supertabular}
\usepackage{booktabs}
\usepackage{multirow}

\begin{document}

\title{ArtCognition: A Multimodal AI Framework \\for Affective State Sensing from Visual \\and Kinematic Drawing Cues}

\author{
Behrad~Binaei-Haghighi,
Nafiseh~Sadat~Sajadi,
Mehrad~Liviyan,
Reyhane~Akhavan~Kharazi, \\
Fatemeh~Amirkhani,
and Behnam~Bahrak

\thanks{Behrad Binaei-Haghighi and Nafiseh Sadat Sajadi contributed equally to this work; Nafiseh Sadat Sajadi and Behnam Bahrak are cocorresponding authors.}%
\thanks{Behrad Binaei-Haghighi, Mehrad Liviyan, and Reyhane Akhavan Kharazi are with the Department of Electrical and Computer Engineering, University of Tehran, Tehran, Iran.}%
\thanks{Nafiseh Sadat Sajadi and Behnam Bahrak are with the Tehran Institute for Advanced Studies, Khatam University, Tehran, Iran.}%
\thanks{Fatemeh Amirkhani is with the Department of Psychology, Allameh Tabataba'i University, Tehran, Iran.}%
}

%\markboth{Journal of \LaTeX\ Class Files,~Vol.~14, No.~8, August~2021}%
%{Shell \MakeLowercase{\textit{et al.}}: A Sample Article Using IEEEtran.cls for IEEE Journals}

%\IEEEpubid{0000--0000/00\$00.00~\copyright~2021 IEEE}

\maketitle
\IEEEpubidadjcol

\begin{abstract}
The objective assessment of human affective and psychological states presents a significant challenge, particularly through non-verbal channels. This paper introduces digital drawing as a rich and underexplored modality for affective sensing. We present a novel multimodal framework, named ArtCognition, for the automated analysis of the House-Tree-Person (HTP) test, a widely used psychological instrument. ArtCognition uniquely fuses two distinct data streams: static visual features from the final artwork, captured by computer vision models, and dynamic behavioral kinematic cues derived from the drawing process itself, such as stroke speed, pauses, and smoothness. To bridge the gap between low-level features and high-level psychological interpretation, we employ a Retrieval-Augmented Generation (RAG) architecture. This grounds the analysis in established psychological knowledge, enhancing explainability and reducing the potential for model hallucination. Our results demonstrate that the fusion of visual and behavioral kinematic cues provides a more nuanced assessment than either modality alone. We show significant correlations between the extracted multimodal features and standardized psychological metrics, validating the framework's potential as a scalable tool to support clinicians. This work contributes a new methodology for non-intrusive affective state assessment and opens new avenues for technology-assisted mental healthcare.
\end{abstract}


% ----------------------------------------------------------------------------------------- Introduction -----------------------------------------------------------------------------------------

\section{Introduction}

\IEEEPARstart{M}{ental} health issues affect a substantial portion of the global population and impose significant personal and societal costs. For instance, in 2019, it was estimated that “one in every eight, or 970 million people in the world, lives with a mental disorder” \cite{who2019}. Despite this high prevalence, accurate diagnosis remains challenging. Current clinical tools, such as the Diagnostic and Statistical Manual of Mental Disorders (DSM) and the International Classification of Diseases (ICD), have limitations, including overlapping criteria, binary categorizations, and neglect of contextual or behavioral factors \cite{kilbourne2018}. This diagnostic gap is particularly problematic in contexts with limited mental health resources, where the lack of specialists delays timely detection and treatment \cite{hanna2018}. Consequently, there is a growing need for innovative tools that can complement conventional diagnostic practices by providing accessible, objective, and scalable assessment methods \cite{balcombe2021}.

Among various clinical assessment techniques, projective drawing tests have a long history of providing insights into human emotions and attitudes, as well as revealing underlying psychodynamics \cite{gleser1963}. The House–Tree–Person drawing test, first proposed by Buck in 1948, remains one of the most widely used projective measures, ranking eighth among commonly applied psychological assessments in an American Psychological Association survey \cite{buck1948htp, camara2000}. The HTP test offers several advantages, including spontaneity, structural complexity, a non-verbal mode of expression, and cultural independence, which collectively enhance its diagnostic utility \cite{smeijsters2006, sheng2019}.

However, traditional drawing assessments face both practical and methodological challenges. Manual scoring is time-consuming and often fails to capture important process-level cues such as the order of drawing objects, number of actions for drawing an object, pauses, and erasing behavior. At the same time, psychology is undergoing rapid digitalization, with an expanding ecosystem of digital tools for mental healthcare \cite{ostermann2021}, enabling AI-driven transformation of projective tests and their analysis. Recent advances in computer vision and deep learning now make it feasible to analyze drawings at scale with greater objectivity and consistency.

Building on these developments, we propose a framework that integrates a digital drawing web app with an AI-assisted interpretation platform to make projective drawing assessments more accessible. The system automatically logs drawing actions, capturing process-level metadata without requiring continuous expert supervision. Moreover, computer vision models detect and classify key components, which are organized into clinically relevant categories using a rule-based psychoanalysis metrics.  Subsequently, a description generator synthesizes the results from the computer vision models and the process-level metadata from the input to create a comprehensive textual description of the drawing. This description is used as a query for a retrieval-augmented generation module which produces explanations grounded in organized psychological knowledge. As a result, using relevant data chunks helps reducing hallucination with combining behavioral logging, object-level analysis, and knowledge-guided reasoning. ArtCognition advances HTP assessment toward a standardized, scalable, and user-friendly workflow.

% ----------------------------------------------------------------------------------------- Related Works -----------------------------------------------------------------------------------------

\section{Related Works}

For interpreting the HTP test, John Buck relied on a combination of structured scoring systems and examiner observations of the drawing process to derive clinical inferences \cite{buck1948htp}, although conventional scoring
protocols are time-intensive and heavily dependent on expert judgment. These limitations motivate automated frameworks to support scalable assessment \cite{guo2023htp}.

Previous research suggests that partial automation of HTP analysis has focused on object detection and rule-based feature extraction. One-stage detectors, such as the YOLO family, provide reliable bounding-box and instance predictions for isolating HTP elements \cite{redmon2016yolo}. Recent architectures, including EfficientNet, Vision Transformer (ViT), ConvNeXt, and Swin Transformer, further improve accuracy for object analysis \cite{tan2019efficientnet, dosovitskiy2020vit, liu2022swin, xie2022convnext}. Lee et al. \cite{lee2023deepdrawing} generated psychological analysis tables from detected elements by computing features such as object proportions and spatial placement, demonstrating potential for more objective and efficient assessment. However, these approaches rely on handcrafted rules and interpret elements in isolation, leaving final report generation to human experts.

Subsequent studies have investigated the use of children’s drawings for emotion classification.
Alshahrani \cite{alshahrani2024drawing} used YOLOv8n-cls to classify drawings into four emotional states (happiness, sadness, anxiety, and anger/aggression), achieving 94\% top-1 accuracy with a compact, mobile-friendly architecture. However, this method is restricted to final drawing and ignores temporal features that could provide richer interpretive signals.

Recent advances in large language models (LLMs), such as GPT-4, have demonstrated strong capabilities in performing tasks that require human-level reasoning, including cognitive psychology challenges. Studies have evaluated GPT-4 on established cognitive datasets such as CommonsenseQA, SuperGLUE, MATH, and HANS, showing high accuracy relative to prior state-of-the-art models. These results suggest that LLMs can integrate contextual information and simulate aspects of human cognitive processes, highlighting their potential to support automated psychological assessment and interpretation~\cite{dhingra2023mind}. 

Two key gaps remain in automated HTP analysis. First, most existing systems focus only on the final drawing, ignoring behavioral metadata which contain valuable information \cite{guo2023htp}. Second, prior approaches often separate recognition from interpretation, with few frameworks integrating object-level detection, behavioral analysis, and psychometric classification in a unified, interpretable pipeline. Addressing these gaps requires end-to-end multimodal models that combine visual, temporal, and contextual features to enable scalable, data-driven, and clinically meaningful psychological assessment.

% ----------------------------------------------------------------------------------------- Methodology -----------------------------------------------------------------------------------------

\section{Methodology}
\begin{figure*}[ht]
    \centering
    \includegraphics[width=0.72\linewidth]{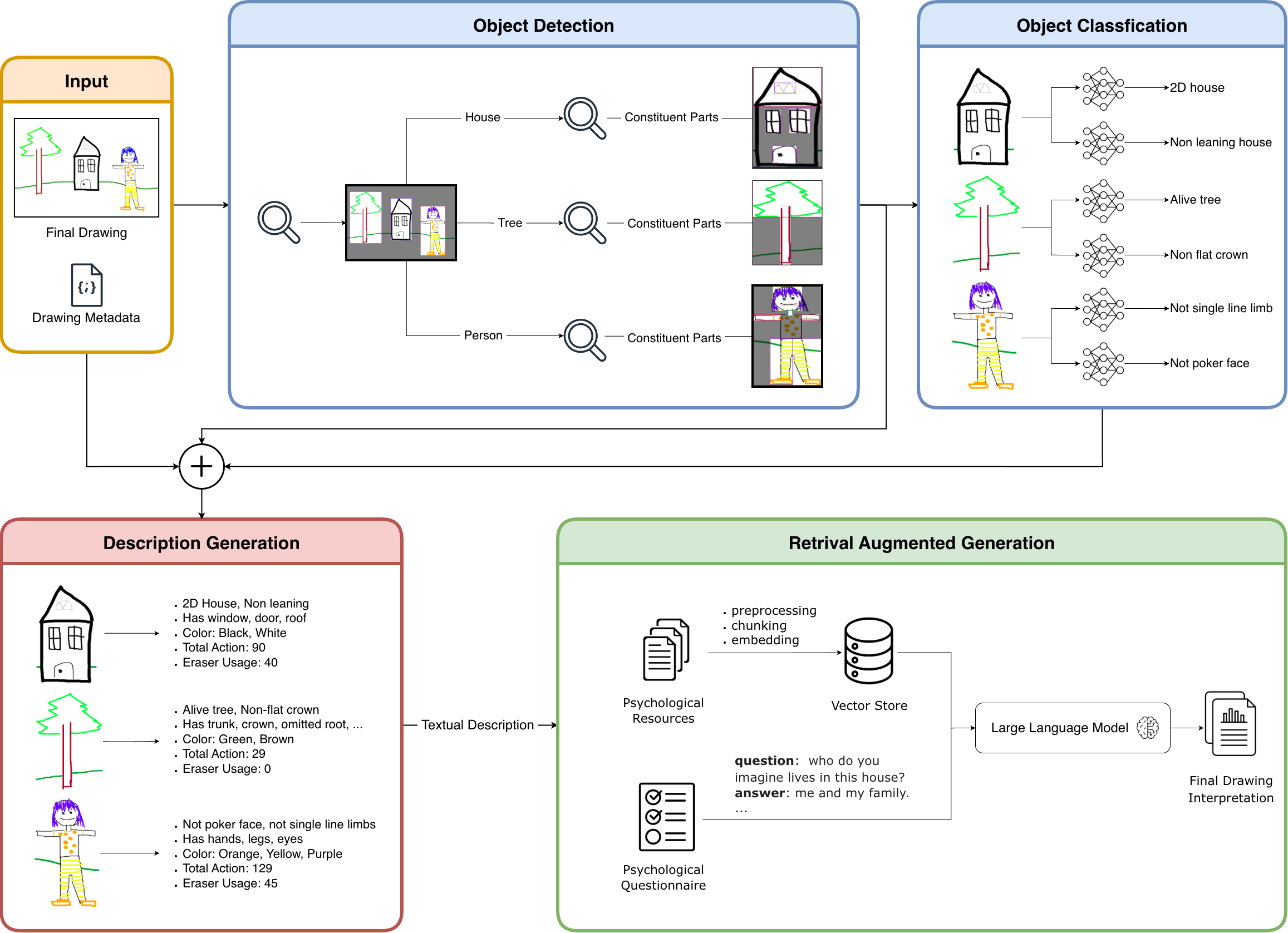}
    \caption{Workflow of ArtCognition}
    \label{fig:workflow}
\end{figure*}

This study addresses the gaps in prior works with a multi-stage approach that integrates image analysis and temporal behavioral data analysis to enable interpretable HTP assessment. First, we construct a dataset of high-resolution drawings paired with fine-grained stroke sequences, capturing both static visual features and behavioral information. Second, we employ a two-stage object detection model to first localize the main house, tree, and person components, followed by a secondary detection stage to identify specific constituent parts within those objects. Third, these detected components are processed through classification models that analyze specific psychological markers, such as a "poker face" or a leaning house. Fourth, we extract behavioral data from the input metadata to track behavioral patterns like eraser usage and stroke actions. Fifth, a description generator synthesizes the object detection results, classification results and metadata into a comprehensive textual summary of the drawing. Finally, this text serves as a query for a retrieval-augmented generation module, which produces final interpretations grounded in expert psychological knowledge. The proposed architecture is illustrated in Figure~\ref{fig:workflow}.

\subsection{Dataset}
To construct the dataset, we use digital drawing to capture fine-grained metadata throughout the drawing process. We collected 146 samples from volunteer participants, each using a custom web-based drawing application designed to record detailed user interactions. The application logs every drawing action with high temporal and spatial resolution, enabling near-exact reconstruction of the drawing process. For each sample, the metadata is stored in JSON format and paired with the final image in PNG format~\cite{htppainter2025}.

\begin{figure}[ht]
    \centering
    % Subfigure 1: Drawing image
    \begin{subfigure}[b]{\linewidth}
        \centering
        \includegraphics[width=0.82\textwidth]{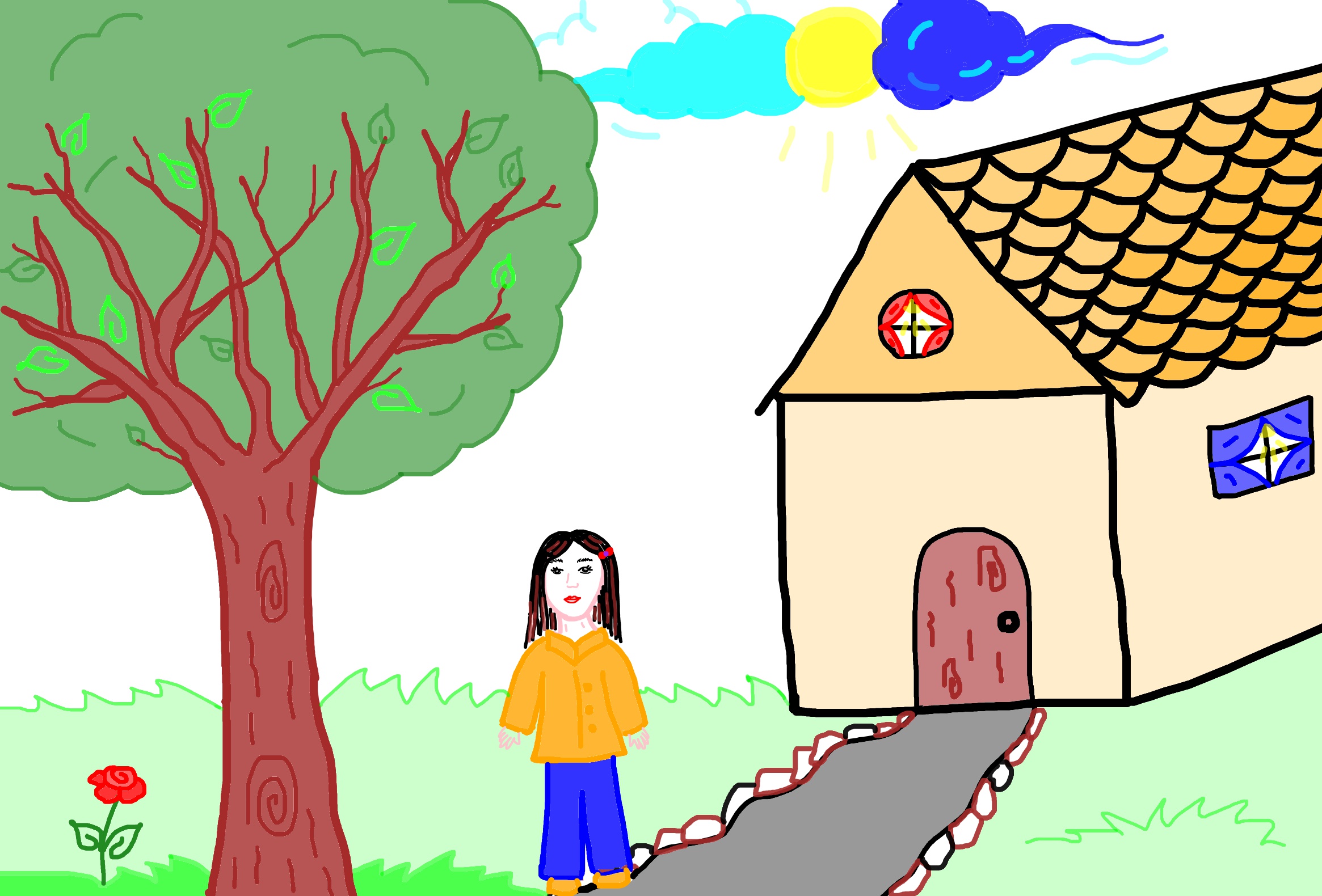}
        \caption{Drawing Artwork}
        \label{fig:drawing}
    \end{subfigure}
    % Subfigure 2: Metadata
    \begin{subfigure}{\linewidth}
        \centering
        %\hspace{10cm}
        \begin{lstlisting}[basicstyle=\ttfamily\scriptsize]
[{
    "order": 1,
    "action_type": "drawLine",
    "color": "#000000",
    "opacity": 1,
    "line_width": 5,
    "timestamp_start": 1751293539626,
    "timestamp_end": 1751293540253,
    "points": [{
        "x": 333.95,
        "y": 102.76,
        "pointerType": "mouse",
        "timestamp": 1751293539626
    }]
}, ...]
        \end{lstlisting}
        \centering
        \caption{Sample of Drawing Metadata}
        \label{fig:metadata}
    \end{subfigure}
    
    \caption{Digital drawing sample and its corresponding metadata.}
    \label{fig:drawing_metadata}
\end{figure}

Each dataset sample consists of the completed drawing along with its corresponding metadata. The metadata records all drawing actions and includes attributes such as drawing order, action type, color, opacity, timestamp, line width, and a sequence of points describing each stroke trajectory. An example of a recorded drawing action is shown in Figure~\ref{fig:drawing_metadata}.

In addition, 21 participants completed a supplementary House–Tree–Person questionnaire (Appendix~\ref{appendix:htp_questionnaire}), developed in collaboration with a domain expert, to provide complementary self-report information and improve the interpretability of the drawings.

\subsection{Object Detection}

The object detection phase uses a two-stage hierarchical approach involving four specialized models based on the YOLOv11 architecture. In the first stage, a main component detector localizes the primary HTP elements: the house, tree, and person. It also identifies environmental features such as clouds, sun, and mountains.

In the second stage, three individual constituent parts detectors, also utilizing YOLOv11, process the cropped images of the house, tree, and person. These models capture fine-grained anatomical and structural details, such as windows and doors in the house, trunks and branches in the tree, and facial features and limbs in the person. This granular detection ensures that all relevant sub-components are isolated for further analysis. These detections directly inform the Description Generation module by confirming the presence or omission of specific elements. Furthermore, the resulting bounding boxes allow for precise spatial measurements that hold significant psychological insights. For instance, house size is classified as \textit{tiny}, \textit{normal}, or \textit{huge}, corresponding to area ratios below $\frac{1}{9}$, between $\frac{1}{9}$ and $\frac{2}{3}$, and above $\frac{2}{3}$ of the drawing, respectively~\cite{takahashi1974htp}.

\subsection{Object Classification}

Once the components and their parts are detected, they are processed through six specialized classification models to identify psychological markers. These models are selected from high-performing architectures, including EfficientNet, ViT, ResNet50, MobileNetV2, ConvNeXt, and Swin Transformer, to ensure high accuracy ~\cite{Sandler2018MobileNetV2IR, He2015DeepRL}.

House Classifiers
Two models analyze the structural characteristics of the house:
\begin{itemize}
\item Leaning House Classifier: Identifies if the house is tilted, which may indicate structural or emotional instability.
\item 2D/3D House Classifier: Determines the perspective of the drawing to assess the level of spatial complexity.
\end{itemize}

Tree Classifiers
Two models evaluate the vitality and shape of the tree:
\begin{itemize}
\item Dead Tree Classifier: Distinguishes between living trees and those depicted as dead or withered.
\item Flattened Crown Classifier: Detects deformations in the tree's crown, such as flattening, which serves as a clinical indicator ~\cite{guo2023htp}.
\end{itemize}

Person Classifiers
The final two models analyze the facial and bodily representation of the person:
\begin{itemize}
\item Poker Face Classifier: Evaluates the facial expression to detect emotional neutrality or a lack of affect ~\cite{guo2023htp}.
\item Single Line Limbs Classifier: Identifies whether the limbs are drawn as simple lines, potentially signaling emotional or developmental constraints ~\cite{meehan1968drawings}.
\end{itemize}

\subsubsection*{Training Process}
All models were trained independently. YOLOv11 was first trained on 117 labeled samples to localize main objects in HTP drawings. Detected objects were then cropped to generate training data for the subsequent classifiers. Once trained, all models were integrated into the pipeline for end-to-end detection and analysis of drawing details.

\subsection{Metadata Analysis}
As illustrated in Figure~\ref{fig:drawing_metadata}, our system captures detailed metadata for each drawing action, including the action type (e.g., drawing a line, using an eraser, using a bucket) and per-point attributes such as coordinates and timing recorded at millisecond resolution, which allow for a precise reconstruction of the drawing\cite{htppainter2025}. More importantly, this data supports advanced analysis that provides quantitative behavioral insights, which are often difficult or impossible for a human or psychologist to measure manually.

\subsubsection*{Stroke Speed}
The Euclidean distance between consecutive points is used to calculate the total stroke path length. This fine-grained measurement allows for the precise calculation of drawing speed for individual strokes and for each detected object (e.g., house, tree, person). The total stroke length $L$ is determined by summing the distances $d_i$ between all consecutive point pairs $(x_i, y_i)$ and $(x_{i+1}, y_{i+1})$. The average stroke speed $v$ is then computed by dividing the total stroke length by the stroke duration $T$. The speed is measured in pixels per second.

\begin{equation}
    d_i = \sqrt{(x_{i+1} - x_i)^2 + (y_{i+1} - y_i)^2},
\end{equation}
\begin{equation}
    L = \sum_{i=1}^{n-1} d_i
\end{equation}
\begin{equation}
    v = \frac{L}{T}
\end{equation}

Drawing speed reflects the ability to control motor, psychomotor, and automatic movements \cite{Rueckriegel2008}. In addition, another study concluded that kinematic parameters such as speed and changes in speed are weaker in children with Developmental Coordination Disorder (DCD) \cite{Abu-Ata2022}.

\subsubsection*{Inter-stroke Interval}

\begin{figure}[htbp]
\centering
\includegraphics[width=0.48\textwidth]{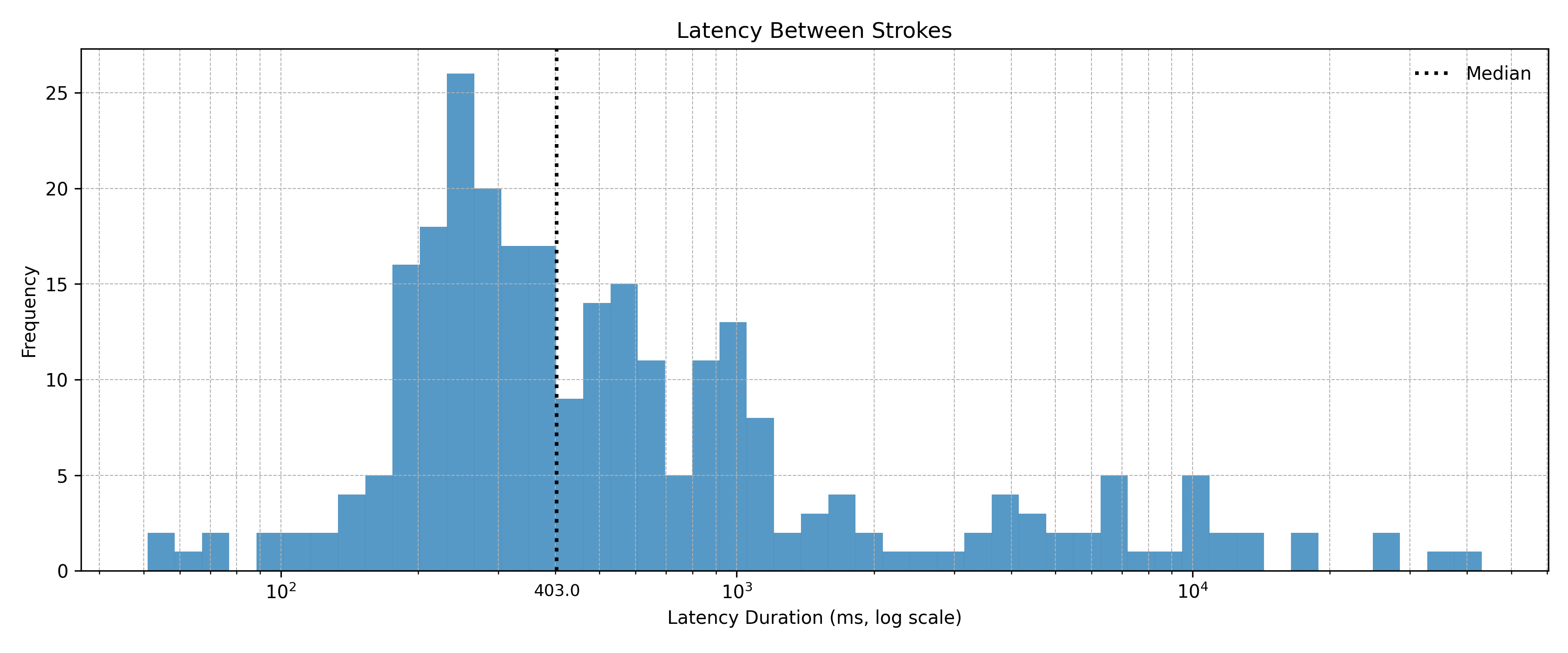}
\caption{Inter-stroke interval throughout the drawing process.}
\label{fig:latency}
\end{figure}

The inter-stroke interval is the time gap between the end of one stroke and the start of the next, extracted directly from timestamped drawing metadata. We compute the total duration of the pause along with statistics such as mean, variance, and distribution across the drawing. These pauses between strokes can be an indicator of cognitive-motor coordination, processing slowness, hesitation, and shifting focus. Moreover, reduced speed was observed particularly in the Lewy Body Dementia (LBD) group, while increased pauses and total durations were observed in both the Alzheimer’s Disease and LBD groups \cite{Yamda2022}. ArtCognition precisely tracks the pauses and speed of the user while drawing, which can enable further research into the behavioral characteristics of specific groups. The distribution of these time gaps and their median value are displayed in the Figure \ref{fig:latency}.
 
\subsubsection*{Stroke Smoothness}
Spectral Arc Length (SPARC) is a common metric used to quantify the smoothness of a line by calculating the arc length of the line’s normalized Fourier magnitude spectrum. The calculation integrates the normalized spectrum up to an adaptive frequency cutoff ($\mathbf{\omega_c}$), where $\mathbf{V(\omega)}$ is the Fourier magnitude spectrum of the line's first derivative (rate of change in pressure) and $\mathbf{\hat{V}(\omega)}$ is the spectrum normalized by its value at zero frequency, $V(0)$. The spectral arc length (SAL) is then defined as: 

\begin{equation}
\begin{array}{l}
    \text{SAL} \triangleq - \int_{0}^{\omega_c} 
    \left[ \left(\frac{1}{\omega_c}\right)^{2} 
    + \left(\frac{d\hat{V}(\omega)}{d\omega}\right)^{2} \right]^{\tfrac{1}{2}} 
    \, d\omega
    \\
    \hat{V}(\omega) = \frac{V(\omega)}{V(0)}
\end{array}
\end{equation}

This methodology, where a value closer to zero indicates greater smoothness, has been applied to analyze kinematic signals  and, notably, to quantify the smoothness of pharyngeal high-resolution manometric curves in swallowing studies ~\cite{scholp2021}.

\subsubsection*{Placement of Drawn Objects}

Neuro-psychological studies show placement patterns change with age and with spatial processing differences ~\cite{Barrett2008}. For example, children tend to draw slightly left of center, with right-handed individuals showing a stronger leftward bias~\cite{picard2014spatialbias}, and similar tendencies appear across cultures. For example, in river basin drawings, children frequently depict rivers flowing from left to right or downwards ~\cite{apostolopoulou2011rivers}.  

To quantify placement, we divide the canvas into a $3 \times 3$ grid, capturing both overall drawing position and individual object locations~\cite{buck2019house}. Each stroke consists of multiple points, and we compute the normalized distribution of points across grid cells:

\begin{equation}
P_{ij} = \frac{n_{ij}}{\sum_{k=1}^{3}\sum_{l=1}^{3} n_{kl}}, \quad i,j \in \{1,2,3\}
\end{equation}

where $n_{ij}$ is the number of points in cell $(i,j)$. This yields a probability distribution $P_{ij}$, allowing comparison across drawings of varying size and density.  

\begin{figure}[htbp]
\centering
\includegraphics[width=0.25\textwidth]{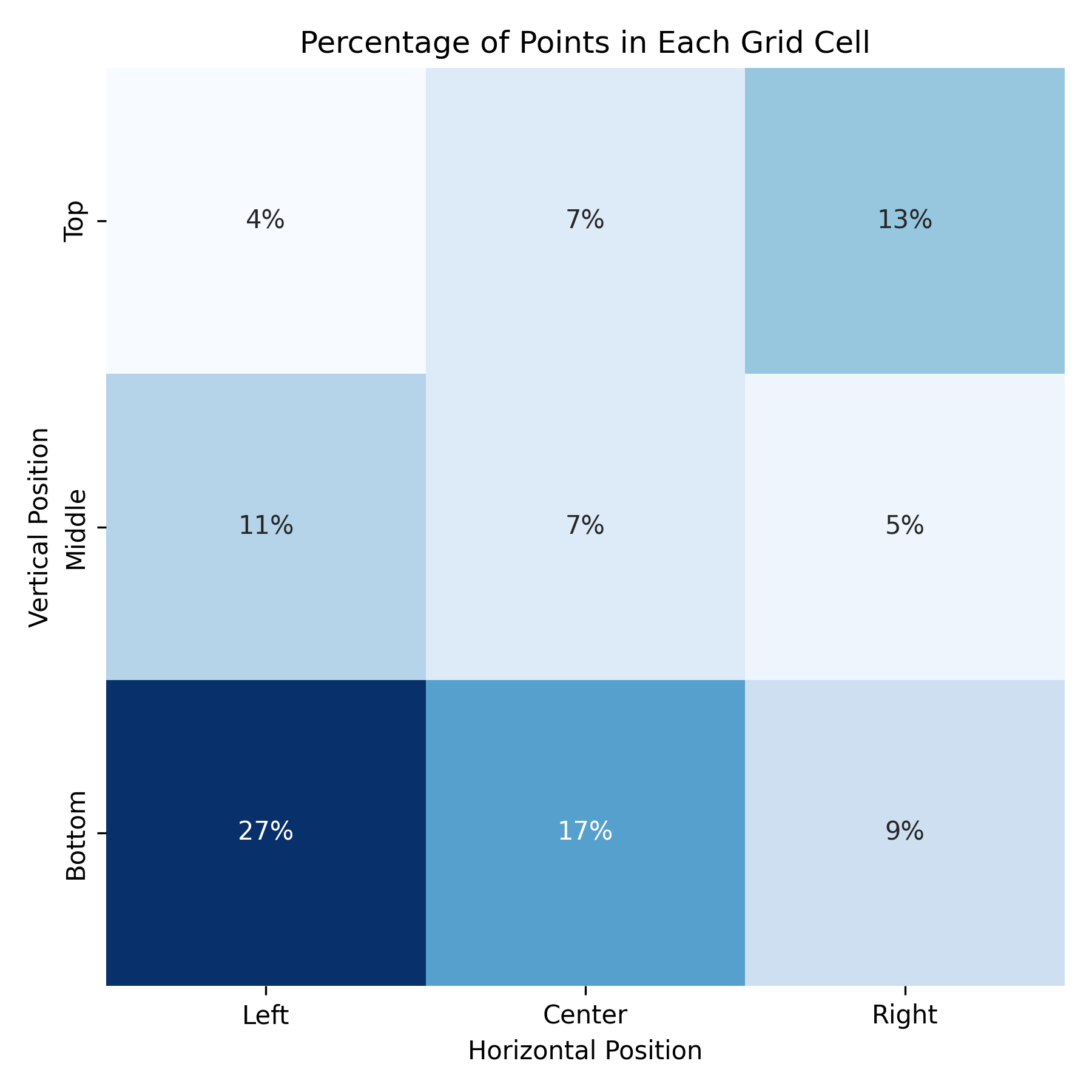}
\caption{Distribution of drawing points over the $3 \times 3$ canvas grid.}
\label{fig:placement}
\end{figure}

Figure~\ref{fig:placement} visualizes the distribution of points across the nine regions, providing a quantitative representation of drawing placement patterns.

\subsubsection*{Eraser Usage Pattern}
High frequency of erasing indicates anxiety, self-doubt, and perfectionism ~\cite{buck2019htp}. On the other hand, a 2020 study compared erasing behaviors in physical and digital drawing environments, finding that erasing occurs more frequently in digital settings, likely due to convenient "undo" functions~\cite{christie2020erase}. These differences did not significantly affect the interpretability of final drawings, suggesting that eraser use reflects compensatory behaviors rather than diagnostic markers.  

In ArtCognition, eraser behavior is analyzed along multiple dimensions. Using bounding boxes from the object detection model, we count erasing events per object to capture user sensitivity to specific elements. We also compute total erasing time and cumulative erased area, providing a behavioral profile of revision strategies and self-monitoring during the drawing process. Generally, redrawn pencil lines evoke a sense of self-criticality and self-doubt ~\cite{leibowitz1999projective}. As a result, our module is capable of providing insights into the user's eraser usage patterns by using metadata. Moreover, with the help of bounding boxes detected by YOLO, our module can determine how many times a user uses the eraser to modify each object.

\subsubsection*{Order of Drawing}

In the HTP test, the sequence in which the examinee chooses to draw the house, tree, or person is considered a behavioral indicator reflecting the individual’s spontaneous focus of attention and emotional priorities~\cite{grothmanat2009}. Moreover, drawing the person first is often associated with heightened self-focus, body-image concerns, or interpersonal sensitivity, particularly when accompanied by detailed corrections or hesitation. Drawing the house first is commonly linked to preoccupation with family relationships, security needs, or domestic concerns. Beginning with the tree is sometimes interpreted as reflecting interest in vitality, strength, or internal resources, especially in children~\cite{handler2013drawings}. Using the bounding boxes detected by YOLO and the action timestamps from the metadata, our module can determine the order and completion time of each drawn object. 

\subsubsection*{Line Width}
The painting interface allows adjustment of brush and eraser width. Across projective drawing research, heavy or thick lines are often associated with heightened tension, aggressive impulses, frustration, or strong emotional activation. In contrast, very light, faint lines may reflect anxiety, insecurity, or low energy~\cite{cooper2010line}.

\begin{figure}[ht]
    \centering
    % Subfigure 1: Drawing image
    \begin{subfigure}[b]{0.48\linewidth}
        \centering
        \includegraphics[width=\textwidth]{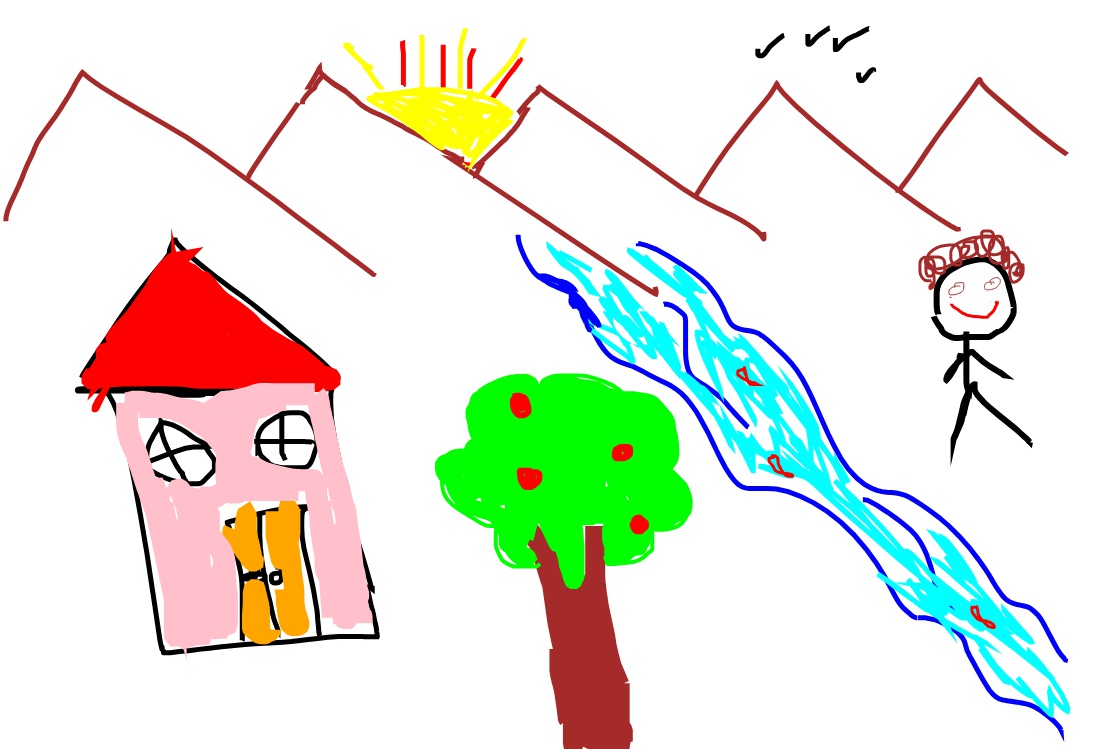}
        \caption{Original Drawing}
        \label{fig:original}
    \end{subfigure}
    \hfill
    % Subfigure 2: Reconstructed image
    \begin{subfigure}[b]{0.48\linewidth}
        \centering
        \includegraphics[width=\textwidth]{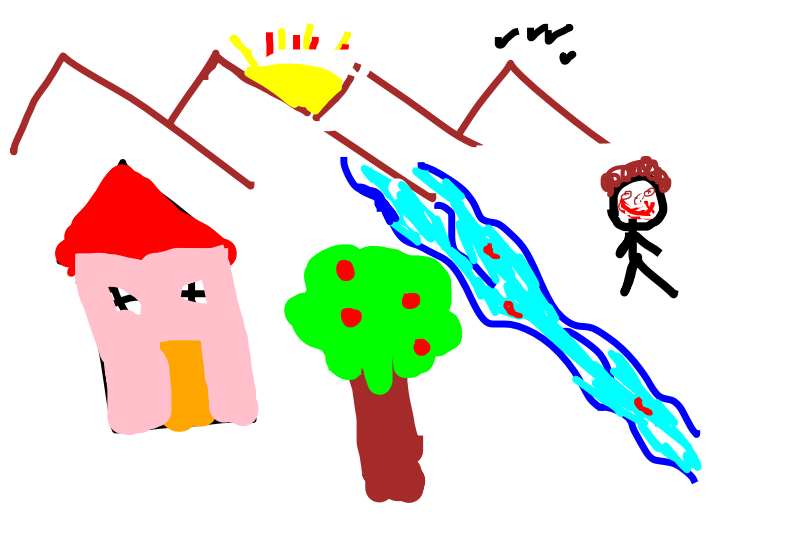}
        \caption{Reconstructed Drawing}
        \label{fig:reconstructed}
    \end{subfigure}
    
    \caption{Original drawing and its reconstruction using metadata.}
    \label{fig:reconstructed_drawing}
\end{figure}

In addition to object detection, we use captured metadata to reconstruct complete drawings. This reconstruction validates the consistency of the logged data and enables a visual replay of the drawing process, facilitating quantitative analyzes that capture the temporal evolution of the sketch rather than only its final outcome. Figure \ref{fig:reconstructed_drawing} shows a comparison between an original drawing and its reconstruction.

\subsection{Description Generation}

The description generation module acts as a synthesis module that converts various data streams into a structured textual representation of the drawing. It integrates three primary sources of information including the object detection results, the classification markers, and the behavioral metadata.

First, the module utilizes the results from the two-stage object detection to identify which elements are present or omitted. For each detected object, it incorporates the findings from the six classification models to describe specific qualitative features, such as a leaning house or a poker face. Second, the extracted data from drawing metadata is further summarized into structured textual descriptions.

\begin{figure}[htbp]
\centering
\includegraphics[width=0.45\textwidth]{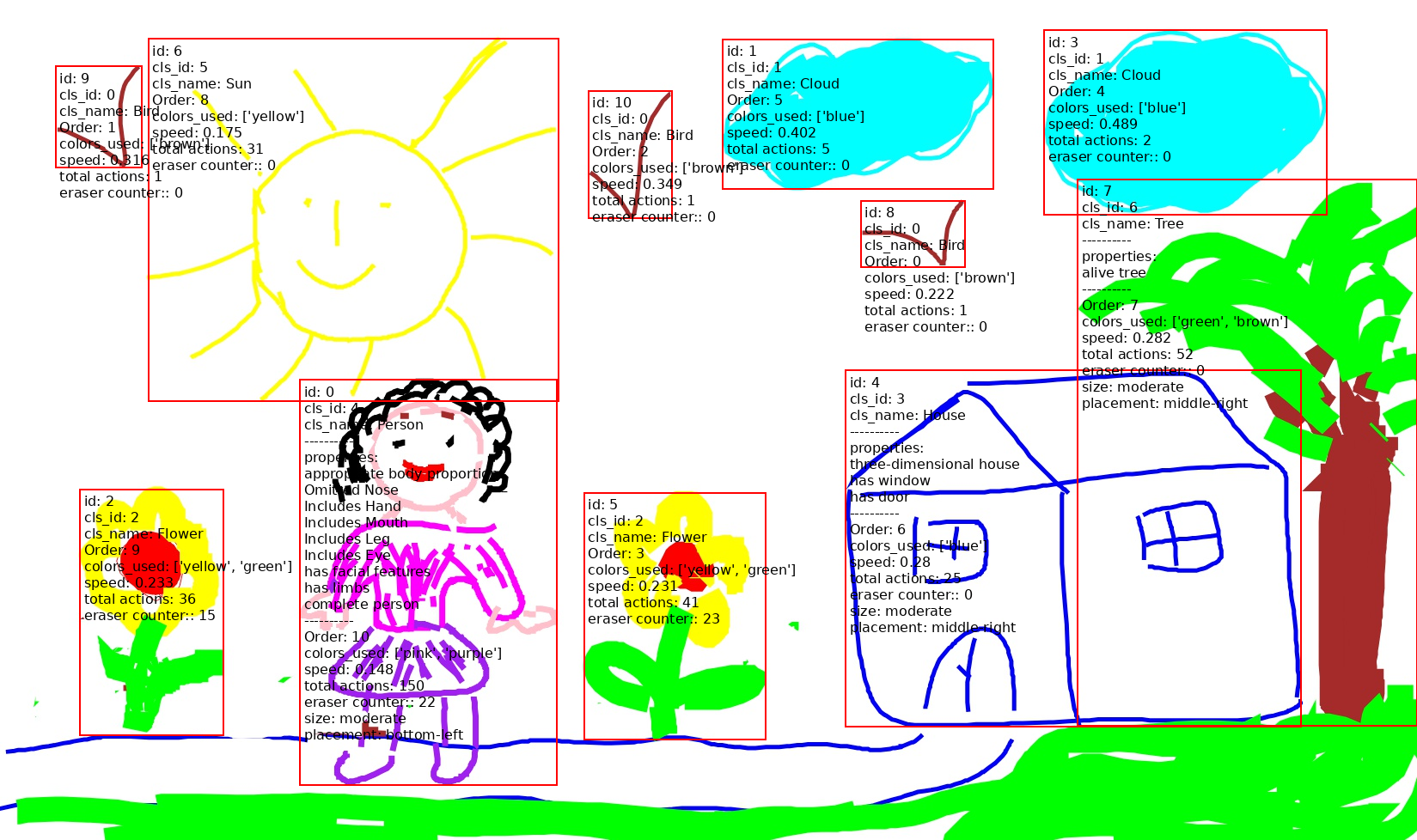}
\caption{Example of how ArtCognition labels objects and records drawing behavior.}
\label{fig:description}
\end{figure}

This structured description serves as the critical bridge between raw visual/behavioral data and the final psychological interpretation. By combining drawn object and the drawing process, the generator provides a comprehensive query for the subsequent RAG phase. The visualization of drawing description is illustrated in Figure~\ref{fig:description}.

\subsection{Psychological Interpretation via RAG Architecture}

The RAG architecture enables the system to transform raw drawing descriptions into expert-level psychological insights. By combining the synthesized data from previous modules with a curated knowledge base, the system generates grounded interpretations rather than isolated observations. This approach ensures that the final output is both evidence-based and aligned with established clinical frameworks.

\subsubsection{Data preprocessing} 
The RAG knowledge base was constructed from authoritative psychological sources, including \textit{Interpreting Projective Drawings: A Self-Psychological Approach}~\cite{leibowitz1999projective} and the study by Guo et al.~\cite{guo2023htp}. Raw text was cleaned by removing non-ASCII characters, followed by stopword removal and lemmatization. To improve retrieval precision, the corpus was categorized according to section headings and subheadings, resulting in standardized data optimized for downstream tasks.

\subsubsection{Data chunking} 
The corpus was segmented using four strategies to evaluate retrieval performance: a) Character-level chunking, fixed-length text segments; b) Recursive character splitting, recursive partitioning along semantic boundaries (paragraphs, sentences) until a size threshold is reached; c) Semantic chunking, grouping sentences by embedding similarity; and d) Semantic clustering, applying K-means on sentence embeddings to form semantically coherent chunks.

\subsubsection{Generating interpretations with an LLM} 
We apply prompt engineering, where retrieved context is provided to the large language model. Conditioning the LLM on domain-specific references grounds the generated interpretations in validated psychological knowledge, mitigates hallucinations, which is a critical concern in clinical AI, and yields interpretable, context-aware reports.

% ----------------------------------------------------------------------------------------- Experimental Results -----------------------------------------------------------------------------------------

\section{Results}

\subsection{Object Detection}

The object detection models were evaluated on a range of drawing elements, including houses, trees, persons, and their constituent parts. Table~\ref{tab:general_objects_single} reports performance across all classes. The main elements such as house, trees, and person achieved high precision and recall, whereas smaller or less frequent elements (e.g., birds) showed lower detection accuracy.

\begin{table}[htbp]
\centering
\caption{YOLOv11 performance on main object detection.}
\label{tab:general_objects_single}
\begin{tabular}{lcccccc}
\toprule
\textbf{Class} & \textbf{\#} & \textbf{P} & \textbf{R} & \textbf{mAP@50} & \textbf{mAP@50-95} \\
\midrule
All & 103 & 0.898 & 0.897 & 0.949 & 0.833 \\
Bird & 4 & 1.000 & 0.710 & 0.768 & 0.421 \\
Cloud & 15 & 0.963 & 0.867 & 0.956 & 0.895 \\
Flower & 3 & 0.567 & 1.000 & 0.995 & 0.908 \\
House & 21 & 0.977 & 1.000 & 0.995 & 0.941 \\
Person & 22 & 0.903 & 0.909 & 0.968 & 0.848 \\
Sun & 6 & 0.949 & 1.000 & 0.995 & 0.949 \\
Tree & 23 & 0.886 & 0.913 & 0.978 & 0.902 \\
Chimney Smoke & 9 & 0.942 & 0.778 & 0.939 & 0.802 \\

\bottomrule
\end{tabular}
\end{table}

The detection results for constituent parts are summarized in Tables~\ref{tab:house_features}–\ref{tab:tree_features}-\ref{tab:person_features}. For houses (Table~\ref{tab:house_features}), windows and roofs achieved strong performance with $\text{mAP@50-95}$ values of $0.92$ and $0.84$, respectively, while doors also showed high accuracy ($0.87$). Chimneys exhibited comparatively lower performance ($\text{mAP@50-95} = 0.61$), likely due to their small size and limited samples. For people (Table~\ref{tab:person_features}), heads were detected most reliably ($\text{mAP@50-95} = 0.73$), followed by eyes, legs, and mouths, whereas finer components such as the nose and neck showed weaker results, reflecting their small spatial extent and higher variability. For trees (Table~\ref{tab:tree_features}), crowns demonstrated robust detection performance ($\text{mAP@50-95} = 0.95$), while trunks and roots achieved moderate accuracy, and branches and fruit yielded lower scores, consistent with their diverse shapes and limited annotations.

\begin{table}[htbp]
\centering
\caption{YOLOv11 performance on house part detection.}
\label{tab:house_features}
\begin{tabular}{lccccc}
\toprule
\textbf{Class} & \textbf{\#} & \textbf{P} & \textbf{R} & \textbf{mAP@50} & \textbf{mAP@50-95} \\
\midrule
All & 83 & 0.965 & 0.836 & 0.940 & 0.808 \\
Chimney & 11 & 0.911 & 0.636 & 0.830 & 0.610 \\
Door & 20 & 0.993 & 0.850 & 0.968 & 0.866 \\
Roof & 16 & 0.987 & 0.938 & 0.983 & 0.837 \\
Window & 36 & 0.971 & 0.920 & 0.980 & 0.917 \\
\bottomrule
\end{tabular}
\end{table}
\begin{table}[htbp]
\centering
\caption{YOLOv11 performance on tree part detection.}
\label{tab:tree_features}
\begin{tabular}{lccccc}
\toprule
\textbf{Class} & \textbf{\#} & \textbf{P} & \textbf{R} & \textbf{mAP@50} & \textbf{mAP@50-95} \\
\midrule
All & 63 & 0.725 & 0.931 & 0.875 & 0.668 \\
Branches & 8 & 0.495 & 0.875 & 0.745 & 0.494 \\
Crown & 23 & 0.968 & 1.000 & 0.995 & 0.950 \\
Fruit & 3 & 0.316 & 1.000 & 0.806 & 0.672 \\
Root & 1 & 0.928 & 1.000 & 0.995 & 0.597 \\
Trunk & 28 & 0.916 & 0.782 & 0.833 & 0.628 \\
\bottomrule
\end{tabular}
\end{table}
\begin{table}[htbp]
\centering
\caption{YOLOv11 performance on person part detection.}
\label{tab:person_features}
\begin{tabular}{lccccc}
\toprule
\textbf{Class} & \textbf{\#} & \textbf{P} & \textbf{R} & \textbf{mAP@50} & \textbf{mAP@50-95} \\
\midrule
All & 182 & 0.822 & 0.762 & 0.825 & 0.601 \\
Eye & 39 & 0.802 & 0.949 & 0.923 & 0.708 \\
Hand & 41 & 0.794 & 0.656 & 0.783 & 0.649 \\
Head & 21 & 1.000 & 0.995 & 0.995 & 0.731 \\
Leg & 39 & 0.892 & 0.795 & 0.884 & 0.686 \\
Mouth & 19 & 0.813 & 0.842 & 0.908 & 0.746 \\
Neck & 14 & 0.906 & 0.688 & 0.827 & 0.333 \\
Nose & 9 & 0.547 & 0.406 & 0.456 & 0.352 \\
\bottomrule
\end{tabular}
\end{table}

In general, object detection models achieve robust performance in main elements and satisfactory detection of constituent parts, providing a reliable foundation for the extraction of downstream features from HTP drawings.

\subsection{Classification Model}

For psychological interpretation, houses, trees, and persons were annotated across six semantic classification tasks (e.g., leaning house vs. non-leaning house). Six models, called ConvNeXt-Base, ViT-B/16, Swin Transformer, EfficientNet-B0, ResNet50, and MobileNetV2, were evaluated using accuracy, precision, recall, and F1-score (Tables~\ref{tab:tree_dead}–\ref{tab:crown_flatten}). The top-performing model for each task was integrated into the pipeline.

% House/Leaning
\begin{table}[htbp]
\centering
\caption{Performance of Leaning House Classifier.}
\label{tab:house_leaning}
\begin{tabular}{lcccc}
\toprule
\textbf{Model} & \textbf{Accuracy} & \textbf{Precision} & \textbf{Recall} & \textbf{F1 Score} \\
\midrule
\textbf{ConvNeXt-Base}     & \textbf{0.952} & \textbf{0.975} & 0.750 & \textbf{0.821} \\
ViT-B/16          & 0.905 & 0.452 & 0.500 & 0.475 \\
Swin Transformer  & 0.857 & 0.639 & 0.697 & 0.659 \\
EfficientNet-B0   & 0.810 & 0.447 & 0.447 & 0.447 \\
ResNet50          & 0.810 & 0.667 & \textbf{0.895} & 0.691 \\
MobileNetV2       & 0.857 & 0.450 & 0.474 & 0.462 \\
\bottomrule
\end{tabular}
\end{table}
% House/Two-dimensional
\begin{table}[htbp]
\centering
\caption{Performance of 2D House Classifier.}
\label{tab:house_2d}
\begin{tabular}{lcccc}
\toprule
\textbf{Model} & \textbf{Accuracy} & \textbf{Precision} & \textbf{Recall} & \textbf{F1 Score} \\
\midrule
ConvNeXt-Base     & 0.619 & 0.683 & 0.653 & 0.611 \\
ViT-B/16          & 0.571 & 0.607 & 0.597 & 0.568 \\
\textbf{Swin Transformer}  & \textbf{0.810} & \textbf{0.846} & \textbf{0.833} & \textbf{0.809} \\
ResNet50          & 0.476 & 0.486 & 0.486 & 0.476 \\
MobileNetV2       & 0.714 & 0.708 & 0.708 & 0.708 \\
EfficientNet-B0   & 0.524 & 0.543 & 0.542 & 0.523 \\
\bottomrule
\end{tabular}
\end{table}
% Tree/DeadTree
\begin{table}[htbp]
\centering
\caption{Performance of Dead Tree Classifier.}
\label{tab:tree_dead}
\begin{tabular}{lcccc}
\toprule
\textbf{Model} & \textbf{Accuracy} & \textbf{Precision} & \textbf{Recall} & \textbf{F1 Score} \\
\midrule
\textbf{ConvNeXt-Base}     & \textbf{0.970} & \textbf{0.978} & \textbf{0.955} & \textbf{0.965} \\
ViT-B/16          & 0.788 & 0.879 & 0.682 & 0.698 \\
Swin Transformer  & 0.848 & 0.907 & 0.773 & 0.802 \\
EfficientNet-B0   & 0.303 & 0.273 & 0.250 & 0.260 \\
ResNet50          & 0.909 & 0.907 & 0.886 & 0.895 \\
MobileNetV2       & 0.788 & 0.879 & 0.682 & 0.698 \\
\bottomrule
\end{tabular}
\end{table}
% Crown/FlattenCrown
\begin{table}[htbp]
\centering
\caption{Performance of Flattened Crown Classifier.}
\label{tab:crown_flatten}
\begin{tabular}{lcccc}
\toprule
\textbf{Model} & \textbf{Accuracy} & \textbf{Precision} & \textbf{Recall} & \textbf{F1 Score} \\
\midrule
ConvNeXt-Base     & 0.818 & 0.474 & 0.429 & 0.450 \\
ViT-B/16          & 0.909 & 0.476 & 0.476 & 0.476 \\
Swin Transformer  & 0.864 & 0.475 & 0.452 & 0.463 \\
\textbf{EfficientNet-B0}   & \textbf{0.955} & \textbf{0.477} & \textbf{0.500} & \textbf{0.488} \\
ResNet50          & 0.909 & 0.476 & 0.476 & 0.476 \\
MobileNetV2       & 0.955 & 0.477 & 0.500 & 0.488 \\
\bottomrule
\end{tabular}
\end{table}
% Face/PokerFace
\begin{table}[htbp]
\centering
\caption{Performance of Poker Face Classifier.}
\label{tab:face_poker}
\begin{tabular}{lcccc}
\toprule
\textbf{Model} & \textbf{Accuracy} & \textbf{Precision} & \textbf{Recall} & \textbf{F1 Score} \\
\midrule
ConvNeXt-Base     & 0.783 & 0.450 & 0.429 & 0.439 \\
\textbf{ViT-B/16 }         & \textbf{0.870} & \textbf{0.455} & \textbf{0.476} & \textbf{0.465} \\
Swin Transformer  & 0.783 & 0.450 & 0.429 & 0.439 \\
ResNet50          & 0.783 & 0.450 & 0.429 & 0.439 \\
MobileNetV2       & 0.565 & 0.433 & 0.310 & 0.361 \\
EfficientNet-B0   & 0.826 & 0.452 & 0.452 & 0.452 \\
\bottomrule
\end{tabular}
\end{table}
% Person/SingleLimbs
\begin{table}[htbp]
\centering
\caption{Performance of Single Line Limb Classifier.}
\label{tab:person_single}
\begin{tabular}{lcccc}
\toprule
\textbf{Model} & \textbf{Accuracy} & \textbf{Precision} & \textbf{Recall} & \textbf{F1 Score} \\
\midrule
\textbf{ConvNeXt-Base}     & \textbf{0.864} & 0.771 & \textbf{0.819} & \textbf{0.790} \\
ViT-B/16          & 0.864 & \textbf{0.781} & 0.722 & 0.745 \\
Swin Transformer  & 0.864 & 0.771 & 0.819 & 0.790 \\
EfficientNet-B0   & 0.864 & 0.771 & 0.819 & 0.790 \\
ResNet50          & 0.818 & 0.675 & 0.597 & 0.614 \\
MobileNetV2       & 0.727 & 0.542 & 0.542 & 0.542 \\
\bottomrule
\end{tabular}
\end{table}

  \subsection{Psychological Interpretation}
To ensure the clinical relevance of our framework, we evaluated the precision of generated description based on drawing and two core components of the RAG pipeline: the retrieval of psychological context and the generation of the final assessment report.

\subsubsection*{Description Generator Performance}
We first evaluate the accuracy of the description generator, which translates vision-based detections and drawing dynamics into structured semantic descriptions used as input to the retrieval module. For each drawing, the generator produces a set of object-level descriptions corresponding to the core HTP elements (house, tree, and person). Each description includes spatial localization (bounding box and placement), categorical attributes (e.g., object type, size, and dimensionality), visual properties (e.g., presence of windows or doors, color usage), and behavioral features derived from the drawing process (e.g., drawing order, stroke speed, and number of actions).

To quantitatively assess description quality, annotators manually reviewed the generated descriptions and counted the number of incorrectly detected or misattributed features for each object. A feature was considered incorrect if it was either falsely detected (e.g., reporting a window when none was present), omitted despite clear visual evidence, or inaccurately described (e.g., incorrect placement or size category). Using these annotations, we compute average precision over the test dataset, defined as the proportion of correctly detected features relative to the total number of predicted features.

Across the evaluation set, the description generator achieves an average precision of 97.57\%, indicating a high level of agreement between the generated descriptions and human annotations. Errors primarily arise in fine-grained visual attributes, such as ambiguous dimensionality (two-dimensional vs.\ three-dimensional representations) and subtle structural elements, while core object identification, spatial placement, and drawing-order features are detected with consistently high accuracy. These results suggest that the generated descriptions provide a reliable and semantically grounded representation of the drawings, forming a stable foundation for subsequent retrieval and psychological interpretation.

\subsubsection*{Retrieval Performance}
Second, we assessed the accuracy of the retriever in fetching relevant psychological interpretations for specific visual descriptions. Ground-truth annotations were established for multiple query sets to measure the semantic alignment of retrieved chunks. As presented in Table~\ref{tab:chunking_cosine_similarity}, we compared various chunking strategies. Semantic chunking yielded the highest cosine similarity ($0.991$), closely followed by K-means semantic clustering ($0.989$). These methods significantly outperformed character-based splitting, indicating that semantically coherent segmentation is crucial for maintaining the integrity of psychological concepts during the retrieval process.

\begin{table}[htbp]
    \centering
    \caption{RAG cosine similarity by chunking strategy.}
    \label{tab:chunking_cosine_similarity}
    \begin{tabular}{l c}
        \toprule
        \textbf{Chunking Strategy} & \textbf{Cosine Similarity} \\
        \midrule
        Character-level chunking & 0.978 \\
        Recursive character text splitting & 0.961 \\
        Semantic chunking & \textbf{0.991} \\
        Semantic clustering with K-means & 0.989 \\
        \bottomrule
    \end{tabular}
\end{table}

\subsubsection*{Generative Quality Assurance and Clinical Validity}
We evaluate the quality and reliability of the generated interpretations through a controlled comparison with two baselines: (1) a rule-based reporting system following Guo et al.~\cite{guo2023htp}, and (2) a standard gemini-2-flash model operating without retrieval augmentation. All outputs were anonymized and reviewed in a blinded setting by licensed clinical psychologists to assess structural accuracy, theoretical consistency, and practical interpretability.

From a computer vision perspective, the proposed framework demonstrates clear advantages in the structured extraction of visual information. The model explicitly decomposes each HTP drawing into semantically meaningful components (house, tree, and person) and analyzes them using specific visual attributes, including spatial configuration (e.g., left-right-center placement), geometric dimensionality (2D vs. 3D representation), color distribution, and feature completeness. This structured decomposition addresses the historical critique of reproducibility in projective testing by replacing impressionistic observation with quantifiable metrics.

Crucially, the integration of drawing features with a retrieval-augmented generation mechanism substantially improves output reliability. While the standard gemini-2-flash baseline often hallucinates by interpreting features that are not physically present or assigning meanings not found in the literature, our framework mitigates this by anchoring its analysis in validated psychological evidence and the participant’s own verbal descriptions. To quantitatively assess this, we measured the hallucination rate, defined as the proportion of interpretive claims referencing unsupported visual or behavioral features. The standard gemini-2-flash baseline exhibits a hallucination rate of approximately 45.72\%. By leveraging the RAG architecture to cross-reference visual detections with established literature, our framework reduces the error rate to zero, as it relies strictly on retrieved reference data; any residual errors originate solely from the object detector or classifier.

Beyond structural accuracy, the generated interpretations reflect a grounded use of high-level theoretical constructs. Visual indicators are not merely described but are consistently linked to psychological states; for instance, spatial orientation and feature omissions are mapped to constructs such as ego development, affective constriction, and perceived internal efficacy.

Furthermore, the model provides clinical value by integrating pictorial data with the participant's self-reported "verbal attributions" (e.g., the perceived age of the tree or affective meaning of the house). The system explicitly identifies areas of convergence and divergence between these modalities. For example, a discrepancy between a self-reported feeling of "calmness" and visual markers of depressive tone is highlighted not as an error, but as a clinically relevant indicator of potential defensive processes or limited emotional awareness.

Finally, the interpretations demonstrate meaningful correspondence with standardized psychometric measures. Visual indicators such as limited color variability, linear limb geometry, and simplified figures were interpreted in a manner consistent with elevated scores on the Beck Anxiety Inventory (BAI) and Beck Depression Inventory (BDI). This alignment suggests that the AI-based analysis is not arbitrary, but possesses convergent validity with established measures of emotional distress.

Overall, the results indicate that ArtCognition produces interpretations that are more structured, reproducible, and evidence-based than ungrounded large language models. While not a substitute for expert clinical judgment, the framework enhances transparency by grounding projective analysis in observable kinematic and visual evidence.

% ----------------------------------------------------------------------------------------- Discussion -----------------------------------------------------------------------------------------

\section{Discussion}

This study presents a reliable and novel pipeline for automated HTP analysis that integrates object detection, classification, kinematic metadata, and RAG architecture. We discuss the effectiveness of this approach, limitations of the dataset and methodology, and ethical considerations of AI-assisted mental health assessment, outlining directions for future work and the potential of combining vision models with behavioral metadata for scalable and interpretable evaluation.

\subsection{Clinical Applicability and Trustworthiness}

Projective drawing tests are widely used in psychology, particularly for assessing children who have difficulty verbalizing emotions~\cite{lee2023deepdrawing, nali2022}. Prior automated HTP interpretation methods often relied on non-standardized scoring systems and handcrafted rules, resulting in inconsistent outputs that can undermine clinician trust~\cite{cai2012, cheng2022, zhou2021, xiang2020, sheng2019}.

ArtCognition addresses these limitations by integrating behavioral metadata with visual features to enhance interpretability. It generates knowledge-grounded reports through the RAG, providing clinicians with a transparent and verifiable basis for analysis while avoiding black-box predictions~\cite{joyce2023, tutun2023}. By automating feature extraction and interpretation, the system reduces the time and effort required for manual scoring~\cite{zeeshan2021}, enabling clinicians to focus on treatment planning and interventions. This efficiency can broaden access to mental health services, while its adaptability and transparent outputs increase clinician trust, supporting broader adoption. Moreover, the design adheres to key principles of trustworthy medical AI, including explainability, and privacy protection~\cite{kim2023}.

\subsection{Limitations}
Despite promising results, this study faces key limitations, including a small dataset, missing behavioral modalities, cultural homogeneity, and constraints of automated interpretation. These challenges point to important directions for future work.

\subsubsection*{Dataset Size}  
The dataset is relatively small, limiting model performance, especially for rare drawing patterns. This scarcity reduces generalization in both detection and classification tasks.

\subsubsection*{Digital Pen Pressure}  
Pen pressure and tilt were not captured, omitting insights linked to neuromotor behavior and limiting the richness of psychological interpretation.

\subsubsection*{Cultural and Demographic Homogeneity}  
All participants were selected from a single cultural background, which may introduce  bias and limit the generalizability of the models to broader populations. 

\subsubsection*{Interpretive Limitations}  
Although high-resolution features of drawings can be quantified with precision, there is limited clinical evidence to support reliable interpretation of many features. Moreover, certain psychological nuances remain difficult to capture computationally, highlighting the continued need for expert supervision.  

\subsection{Ethical Considerations}  
Automated psychological assessment raises important concerns regarding privacy and responsible use. Addressing these issues is critical for the safe and ethical deployment of digital mental health applications.

% ----------------------------------------------------------------------------------------- Conclusion -----------------------------------------------------------------------------------------

\section{Conclusion}

We present a state-of-the-art AI pipeline for automated analysis of digital HTP drawings, integrating object detection, classification, behavioral metadata, and a RAG architecture for psychological interpretation.
 By combining visual features, kinematic data, and external psychological knowledge, the system provides interpretable and scalable assessments that go beyond traditional manual scoring. Experimental results demonstrate high detection and classification accuracy, in addition metadata analysis captures nuanced behavioral indicators such as drawing speed, line smoothness, and drawing order.

Key limitations include a small, culturally homogeneous dataset, the absence of pen pressure data, and the inherent challenge of quantifying complex psychological states. Future work will expand the dataset, incorporate additional neuromotor modalities, and validate the approach across diverse populations. Overall, this study demonstrates the potential of AI-assisted digital drawing analysis as a complementary tool for psychological assessment, offering efficiency, interpretability, and evidence-based insights.

% ----------------------------------------------------------------------------------------- Acknowledgments -----------------------------------------------------------------------------------------

\section*{Acknowledgments}
The authors would like to thank Shayan Talebiani for his contributions to data labeling and for implementing the metadata-based module used to compute the number of actions associated with each object.

% ----------------------------------------------------------------------------------------- Appendix -----------------------------------------------------------------------------------------

\appendices
\section{Dataset}
\label{appendix:dataset}
This appendix presents comprehensive dataset statistics, detailing the frequency of object classes and their constituent parts. These metrics are provided to facilitate reproducibility and offer essential context for the experimental results and analyses discussed in the main text. Table \ref{tab:dist_main_obj_class} presents the frequency of the main object categories in the dataset, showing how often each high-level object (e.g., house, tree, person) appears. It provides an overview of the dataset’s overall composition and helps illustrate class balance.

\begin{table}[H]
	\centering
	\caption{Distribution of main object classes in the dataset.}
	\label{tab:dist_main_obj_class}
	\scriptsize
	\begin{tabular}{lc}
	\toprule
	\textbf{Class} & \textbf{\# Instances} \\
	\midrule
		Bird & 66 \\
		Cloud & 112 \\
		House & 147 \\
		Tree & 181 \\
		Person & 156 \\
		Flower & 46 \\
		Mountain & 14 \\
		Sun & 63 \\
		Chimney Smoke & 45 \\
	\bottomrule
	\end{tabular}
\end{table}

\begin{table}[H]
    \centering
    \caption{Distribution of house, tree, and person object classes in the dataset.}
    \label{tab:combined_distribution}
    \scriptsize
    
    % --- Table (a): House ---
    \begin{subtable}[t]{0.3\linewidth}
        \centering
        \caption{House}
        \begin{tabular}[t]{lc}
            \toprule
            \textbf{Class} & \textbf{\# Inst.} \\
            \midrule
            Door    & 135 \\
            Window  & 249 \\
            Roof    & 107 \\
            Chimney & 58 \\
            \rule{0pt}{0.9em} & \\ % Empty row 1
            \rule{0pt}{0.9em} & \\ % Empty row 2
            \rule{0pt}{0.9em} & \\ % Empty row 3
            \bottomrule
        \end{tabular}
    \end{subtable}
    \hfill
    % --- Table (b): Tree ---
    \begin{subtable}[t]{0.3\linewidth}
        \centering
        \caption{Tree}
        \begin{tabular}[t]{lc}
            \toprule
            \textbf{Class} & \textbf{\# Inst.} \\
            \midrule
            Branches & 40 \\
            Crown    & 158 \\
            Fruit    & 144 \\
            Root     & 21 \\
            Trunk    & 180 \\
            \rule{0pt}{0.9em} & \\ % Empty row 1
            \rule{0pt}{0.9em} & \\ % Empty row 2
            \bottomrule
        \end{tabular}
    \end{subtable}
    \hfill
    % --- Table (c): Person ---
    \begin{subtable}[t]{0.3\linewidth}
        \centering
        \caption{Person}
        \begin{tabular}[t]{lc}
            \toprule
            \textbf{Class} & \textbf{\# Inst.} \\
            \midrule
            Eye   & 213 \\
            Hand  & 266 \\
            Head  & 134 \\
            Leg   & 258 \\
            Mouth & 105 \\
            Neck  & 74 \\
            Nose  & 37 \\
            \bottomrule
        \end{tabular}
    \end{subtable}
\end{table}

\section{House--Tree--Person Questionnaire}
\label{appendix:htp_questionnaire}
This questionnaire provides additional contextual information about participants’ interpretations of their drawings. It was designed by a clinical expert to capture subjective perceptions related to the house, tree, and person figures.

\subsection{Participant Information}
\noindent Name: \underline{\hspace{1cm}} \hfill Age: \underline{\hspace{1cm}} \hfill Gender: \underline{\hspace{1cm}}

\subsection{Questions}
\begin{enumerate}
    \item Who do you imagine lives in this house?
    \item What feelings does this house give you?

    \item How old do you think the tree is?
    \item Is the tree alive or dead?
    \item Which season of the year do you think it is?

    \item Does this image remind you of anyone?
    \item How old do you think this person is?
    \item What do you think this person does?
    \item What might this person be thinking?
    \item What do you think this person feels?
\end{enumerate}

\section{Sample of drawing interpretation using LLM}
\label{appendix:sample_interpretation}

\begin{figure}[htbp]
\centering
\includegraphics[width=0.45\textwidth]{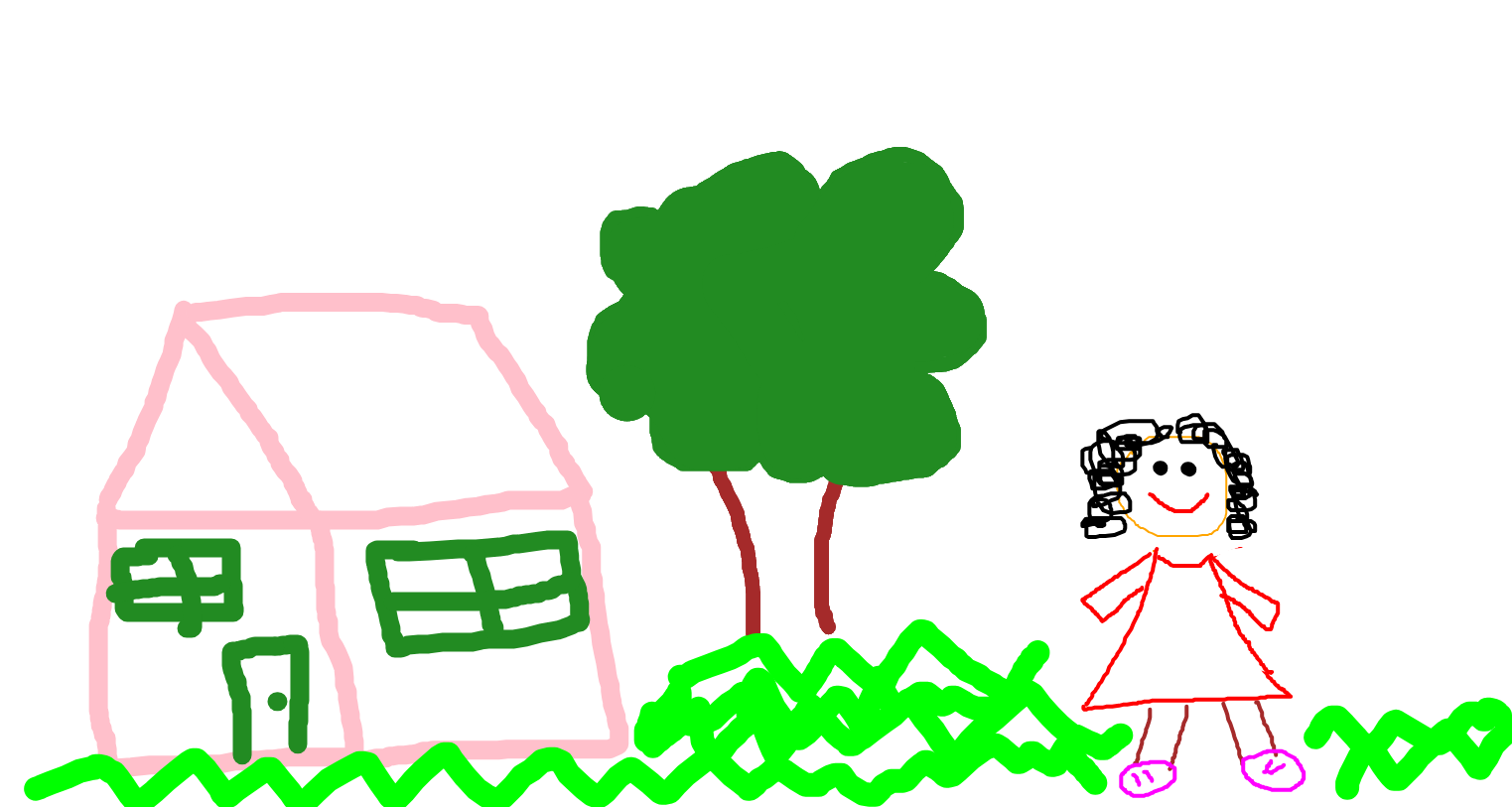}
\caption{Sample drawing used to compare vanilla and ArtCognition interpretation methods.}
\label{fig:input_image}
\end{figure}

Figure \ref{fig:input_image} shows the drawing used as the primary input for sample analysis. While the vanilla model only looks at the static image, ArtCognition integrates various data layers to generate a more comprehensive evaluation. The following sections compare the basic interpretation with the advanced output provided by ArtCognition.

\subsection{Vanilla Interpretation}
\label{subsec:vanilla}

\subsubsection{Overall Impression}
The drawing presents a simple, colorful scene of a house, tree, and person, suggesting a focus on basic elements of life and environment. The bright colors and smiling face indicate a generally positive affect. The composition appears somewhat spread out across the page, potentially suggesting a need for more structure or boundaries.

\subsubsection{House}
\begin{itemize}
\item   Color: The pink color of the house could represent warmth, affection, or a feminine association.
\item   Shape: The house is drawn in a simple, conventional form, suggesting adherence to norms and a desire for stability.
\item   Windows and Door: The presence of windows and a door indicates openness to social interaction and a desire to connect with the outside world. The windows are large and clearly defined, further emphasizing this openness.
\item   Foundation: The house appears to be grounded by a wavy green line, possibly representing the environment and a sense of connection to nature.
\end{itemize}

\subsubsection{Tree}
\begin{itemize}
\item   Size and Shape: The tree is large and has a full, rounded crown, suggesting a strong sense of growth, vitality, and emotional expression.
\item   Trunk: The trunk is depicted with two lines, which could indicate strength and stability.
\item   Placement: The tree is positioned between the house and the person, potentially symbolizing a bridge between the home environment and the individual's self-expression.
\end{itemize}

\subsubsection{Person}
\begin{itemize}
\item   Facial Expression: The smiling face is a clear indicator of happiness and a positive outlook.
\item   Hair: The curly hair may represent energy, creativity, or a desire for attention.
\item   Arms and Legs: The arms are extended, suggesting a willingness to engage with the world. The legs are simple, indicating a sense of groundedness.
\item   Clothing: The dress could represent femininity or a sense of self-presentation.
\item   Shoes: The pink shoes add a touch of whimsy and individuality.
\end{itemize}

\subsubsection{Color Usage}
The use of bright colors like pink, green, and red suggests optimism, energy, and a desire for stimulation.

\subsubsection{Placement on the Page}
The drawing is positioned in the lower half of the page, which could indicate a sense of security or a need for support.

\subsubsection{Limitations}
It's crucial to remember that this is a basic interpretation based solely on the drawing. A comprehensive analysis would require additional information about the artist's background, experiences, and emotional state. This interpretation should be considered a starting point for further exploration, not a definitive diagnosis.

\subsection{ArtCognition's Interpretation}
\label{subsec:rag}

\subsubsection{Overall Impression}
The drawing, featuring a house, tree, and person, is centrally placed, suggesting a feeling of security and a capacity to relate to the present moment. However, the exact center placement can also evoke a need for balance and rigidity. The simplified nature of the drawing, coupled with the lack of additional decoration, may indicate low mental motivation or a tendency towards avoidance and retreat. The use of normal coloring suggests the drawer can express a wide range of subjective experience.

\subsubsection{House}
\begin{itemize}
\item   General: The three-dimensional house signifies an attempt to represent depth. The drawer identifies with the house as her childhood home, associating it with happiness and a lack of sadness, indicating a potentially positive view of her past.
\item   Door: The presence of a door suggests a degree of accessibility, but without a doorknob, it conveys uncertainty or ambivalence about letting others gain access.
\item   Windows: The presence of windows indicates a capacity for relatedness and interaction with the surroundings.
\item   Color: The pink color of the house is not directly referenced in the provided text, thus hard to provide interpretation.
\end{itemize}

\subsubsection{Tree}
\begin{itemize}
\item   General: As a tree is alive, and depicts summer, it suggests a sense of vitality and aliveness. The drawer sees the tree as 20 years old, while she is 29, and there is a diminished sense of felt maturity.
\item   Color: The green color evokes aliveness and expansiveness.
\end{itemize}

\subsubsection{Person}
\begin{itemize}
\item   General: The complete person with appropriate body proportions suggests an intact sense of self-cohesion. The moderate size of the person indicates an optimal sense of adequacy. The figure reminds the drawer of herself as a little girl, reinforcing the connection to childhood.
\item   Facial Features: The presence of facial features indicates a concern about appearance and how one presents oneself to others. However, the omission of the nose suggests a serious disturbance in how one appears to others and receives affective stimuli.
\item   Limbs: The inclusion of limbs (legs and hands) indicates a capacity for maneuverability and connection with the environment.
\item   Color: The use of pink on the person again has no direct meaning, but a general observation can be made: chromatic drawings allows for a wider range of expression.
\end{itemize}

\subsubsection{Color}
\begin{itemize}
\item   Green: The use of green suggests a sense of aliveness and expansiveness.
\item   White: The white background alludes a loss of self-awareness and psychological defenses.
\item   Pink: Color is not referenced in the provided text.
\end{itemize}

\subsubsection{Placement}
The central page placement indicates normality.

\subsubsection{Summary}
The drawing suggests a connection to the drawer's childhood, associating it with happiness. There is a sense of vitality and a desire for connection, but also some ambivalence about accessibility and a tendency toward withdrawal, possibly as a defense. The omission of the nose and the simplified style suggest some underlying anxieties related to self-perception and interaction with the world. The central placement and moderate size of the figures indicate a basic sense of security and adequacy, although the need for balance and control may be a factor.

\subsection{Accuracy Score Computation}
To quantify interpretive hallucinations, we measure an accuracy score based on expert validation. For each generated interpretation, individual claims are decomposed into atomic interpretation units. Each unit is labeled as \emph{correct} or \emph{incorrect} according to domain-expert (psychologist) feedback and established HTP methodological rules. Omissions are not penalized unless replaced by an incorrect inference. For example, in a representative case, 13 interpretation units were identified, of which 5 were judged correct and 8 were judged incorrect due to unsupported symbolic or affective inferences. The final accuracy score is approximately $38.5\%$, where lower value indicate a higher propensity for hallucination in vanilla interpretation.

\section{House--Tree--Person Drawing Characteristics}
\label{appendix:htp_comparison}

Table~\ref{tab:drawing_psychology} summarizes detailed information on the drawing characteristics analyzed in the House--Tree--Person test and their associated psychological interpretations. We use this table to compare our method with the study by Guo et al.~\cite{guo2023htp}, highlighting the observable drawing features and the corresponding inferred psychological constructs.

\begin{table}[htpb]
    \centering
    \scriptsize
    \setlength{\tabcolsep}{3pt} % reduce horizontal padding
    \caption{Drawing characteristics and their associated psychological interpretations.}
    \label{tab:drawing_psychology}
    \begin{tabular}{p{4.5cm}p{3.5cm}} % swapped widths
        \toprule
        \textbf{Drawing Characteristics} & \textbf{Indicates Meaning} \\
        \midrule
        Excessive separation among items; Omitted house, tree, or person; No door / window; Loss of facial features / poker face; Complete or partial loss of limbs; Incomplete person; Left page placement / upper-left corner placement; Color: white
        & Loss of self-awareness and psychological defenses \\
        \midrule
        Leaning house; dead Tree; flattened crown; Inappropriate body proportions; Fist; High or right page placement; Colors: purple, brown
        & Psychological conflict and sense of unreality \\
        \midrule
        Smoking chimney; Roots; Colors: yellow, purple; Top edge page placement
        & Nervousness, sensitivity, and irritability \\
        \midrule
        No additional decoration; Simplified drawing; Small drawing size; Emphasis on straight lines; Very small house / tree / person; Two-dimensional house; Single line limbs; Absence of color; Low page placement; Color: white; Faint lines; Left page placement
        & Low mental motivation, avoidance, and retreat \\
        \midrule
        Left page placement; Low page placement; Colors: brown, white; Upper-left corner placement
        & Regression \\
        \midrule
        Central page placement; Colors: orange, green, blue
        & Normality \\
        \midrule
        Low page placement; Small drawing size; Very faint lines; Color: white
        & Depression, emptiness \\
        \midrule
        Low page placement; Small drawing size; Color: brown; Left/top edge page placement
        & Insecurity \\
        \midrule
        Large drawing size; Heavy/thick lines
        & Aggression \\
        \midrule
        Left page placement; Small drawing size; Very faint lines; Low page placement; Color: green
        & Self-esteem, childish \\
        \midrule
        Bottom edge of paper
        & Need for external support, dependence \\
        \midrule
        Side edge of paper; Large drawing size; Excessive use of color
        & Environmental restriction, pressure \\
        \midrule
        Large drawing size; Colors: red, orange; High page placement
        & Heightened vitality, energy, manic states \\
        \bottomrule
    \end{tabular}
\end{table}

% ----------------------------------------------------------------------------------------- References -----------------------------------------------------------------------------------------

\bibliographystyle{IEEEtran}
\bibliography{references}

\end{document}